# Separating Topology and Geometry in Space Planning

**Benachir Medjdoub[*] and Bernard Yannou[**]**

[*]The Martin Centre, University of Cambridge
6 Chaucer Road, CB2 2EB, Cambridge, UK
Phone: (01223)331714
Fax: (01223)331701
bm230@cam.ac.uk

[**]    Laboratoire Productique Logistique, Ecole Centrale Paris,
Grande Voie des Vignes, 92295 Chatenay Malabry, France
Phone: (33) 1 41 13 16 05 –
Fax: (33) 1 41 13 12 72
yannou@pl.ecp.fr

We are here dealing with the problem of space layout planning. We present an architectural conceptual CAD approach. Starting with design specifications in terms of constraints over spaces, a specific enumeration heuristics leads to a complete set of consistent conceptual design solutions named topological solutions. These topological solutions which do not presume any precise definitive dimension correspond to the sketching step that an architect carries out from the Design specifications on a preliminary design phase in architecture.

## 1. INTRODUCTION

Many architects are confronted on a daily basis with the problem of space layout planning, i.e. the best space arrangement with regards to objective requirements. Objective requirements are expressed by constraints:

- Dimensional constraints: over one space such as constraints on surface, length or width, or space orientation.
- Topological constraints: over a couple of spaces such as adjacency, adjacency to the perimeter of the building, non-adjacency, proximity.

Currently, architects solve these placement problems "by hand". Traditionally, starting from specification constraints, they start by drawing some sketches which represent space planning principles or topologically feasible solutions with no precise geometrical dimensions. This is the sketch stage. Next, geometrical dimensioning is more dependent upon objective requirements (good space proportion, minimum surface area required…). For this automatic or manual geometrical stage, architects may use parametric or variational CAD softwares. These programs allow the architects to directly handle a parameterized space planning.

At the present time, the main weaknesses of this methodology are in the sketch research stage. On the one hand, an architect may omit some sketches. On the other hand, some sketches, found by the architect, which are apparently topologically sound, turn out, in fact, to be inconsistent solutions, when trying to evaluate space dimensions.

Many attempts of space layout planning in architecture have used expert systems (André, 1986; Flemming, 1988). These approaches present many disadvantages: we are never sure of the completeness and the consistency, we are never sure of obtaining the global optimum, and reply times are long.

Another recent approach, the evolutionary approach (Damski and Gero, 1997; Jo and Gero, 1997; Gero and Kazakov, 1998; Rosenman, 1996) is an optimization process which deals with practical problems (up to 20 spaces and several floors) but results are sub-optimal solutions.

It has been shown that constraint programming techniques bring, a great flexibility in the constraint utilization since the constraint definition is separated from resolution algorithms, as well as highly combinatorial problems as is the case for optimal placement (Aggoun and Beldiceanu, 1992; Charman, 1994; Baykan and Fox, 1991).

All these approaches enumerate all the placement solutions. Then, two quasi-equivalent solutions, where only a partition is translated by a *module*[1] are considered as two different *geometrical solutions* (see Figure 1). It is clear that, in preliminary design, it is useless to discriminate between two geometrical close solutions, as this provokes an explosion of solutions (typically several thousands or millions) which cannot be apprehended in their globality by the architect. In addition, they are too precise at this design stage. Conceptual designs are more judicious in a first stage, they can be compared to architects' sketches in this primary research of placement principles.

Several approaches (Mitchell et al, 1976, Schwarz et al 1994), based on a graph-theoretical model, have already introduced the topological level as a part of the computational process. Contrarily to our approach, the topological level does not allow any initial domain reduction of the variables. This fact makes impossible to evaluate or graphically represent the topological solutions. The evaluation and the graphical representation of the solutions are done at the geometrical level.

---

1 Architects define a module as the distance increment for the space dimensions (width, length) and the grid spacing. The grid is the grid of columns, beams and load-bearing walls.

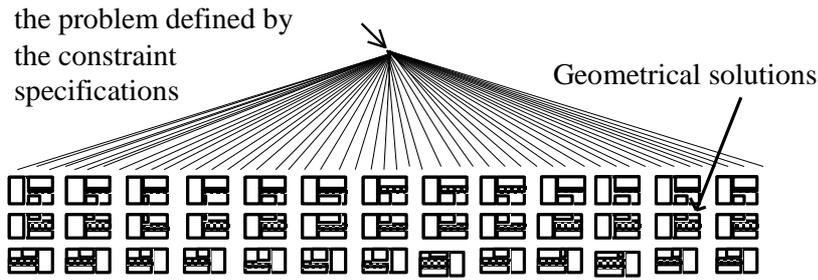

**Figure 1** Conventional approaches in constraint-based space layout planning. An exhaustive enumeration of geometrical solutions is performed.

Our approach and its implementation within the ARCHiPLAN prototype is based on a constraint programming approach which importantly avoids the inherent combinatorial complexity for practical space layout problems. Moreover, we propose to get closer to natural architect's design processes in considering a primary solution level of *topological solutions*. These topological solutions must respect the specification constraints of the design problem and they must lead to consistent geometrical placement solutions (see Figure 2). For that purpose, we have proposed a new definition of a topological solution and we have developed a specific topological enumeration heuristics.

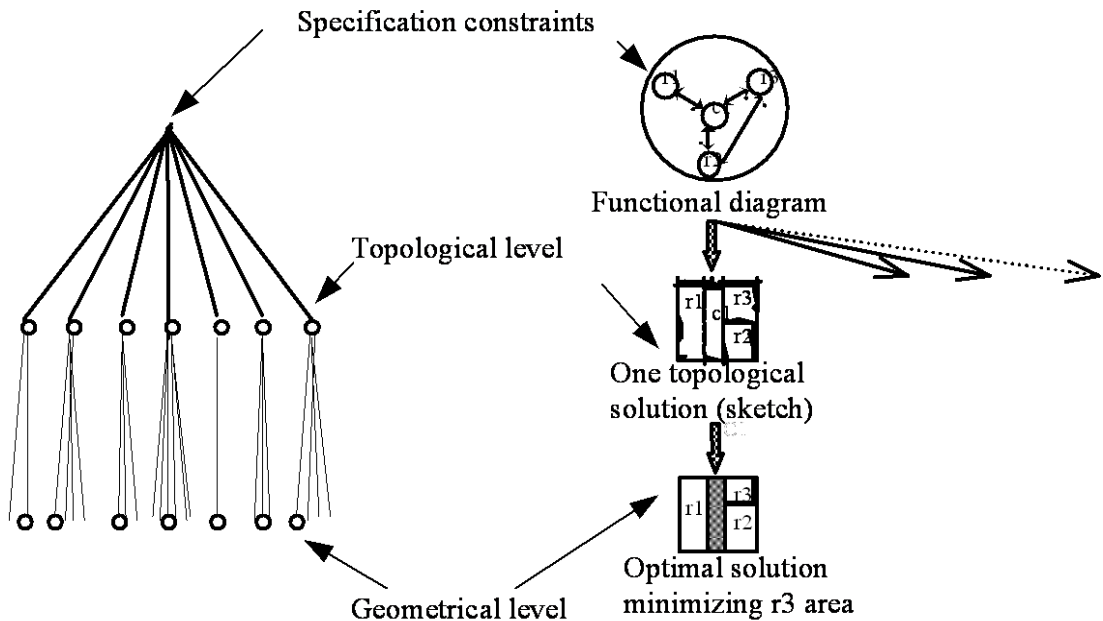

**Figure 2** Solution levels in ARCHiPLAN: topological and geometrical. For each topological solution, only the best corresponding geometrical solution is calculated or few are tested.

Our topological solution turns out to be an equivalence class of geometrical solutions respecting the same conditions of relative orientation (north, south, east, west) between all the pairs of spaces. Thus, two *topologically different* solutions, are differentiated by at least one different adjacency. We noticed that such a *topological solution* representation corresponds to a sketch drawing, i.e. a sketch made by the architect in the preliminary design. The advantage of the *topological solution level* is the low number of existing solutions (typically less than one hundred), a number that can be easily apprehended by the architects. Architects are now able to have a global view of all the design alternatives ; they will then only study in detail a small number of topologies which correspond to their aesthetic appreciation, as in practice. Anyway, thank to the optimization, a geometrical step determines the best geometrical placement solution for each topological solution from a set of user-defined criteria. On the one hand, optimization leads to geometrical solutions minimizing or maximizing criteria such as wall-length or some surface area, these criteria are useful for architects. On the other hand, optimization limits the number of solutions. This result turns out to be a determining decision-making tool because it allows comparizon between topological solutions in terms of their realizability.

In section 2, we briefly present the architectural model. In section 3, we present our constraint model. The algorithm of topological solution enumeration is presented in section 4 and the geometrical solution optimization is presented in section 5. Before concluding, we present a case study in section 6.

## 2. MODEL OF ARCHITECTURAL SPACE REPRESENTATION

Our knowledge model holds the main architectural elements corresponding to empty spaces, i.e. which are neither not structural elements (walls, beams, windows, etc.) nor furniture. Each defined class is characterized by attributes and class constraints. After presenting the generic *space* class, we describe the two main classes of our architectural space model: *room* class and *stair* class.

### 2.1 Space class

As we deal with orthogonal geometry, we call *space* an isothetic rectangle (see Figure 3), which is representative of an important part of architectural problems. This class is characterized by an identifier, two reference points *(x1, y1)* and *(x2, y2)* (at the opposite of the rectangle), a length *L*, a width *W* and a surface area *S*. All these attributes, except the identifier, are integer constrained variables. We used an *arc-consistency on integers* constraint programming technique which explains the need for a *distance increment* ; but this is not too limitative as architects are used to reasoning with dimensional *modules*. L-shape and T-shape are allowed, and they correspond to two adjacent spaces with a minimum contact length.

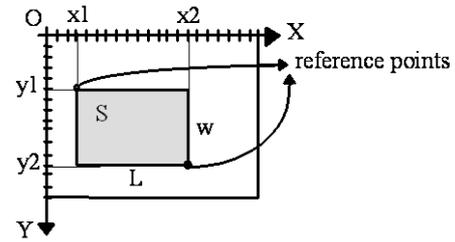

**Figure 3** Space class geometrical representation.

The three following class constraints have been defined so as to ensure the geometrical consistency of the space classes:

• (c1) $x2 = x1 + L$

• (c2) $y2 = y1 + W$

• (c3) $S = L \times W$

A modification of a variable domain composing the constraint (c1), (c2) or (c3) entails the modification of variable domains of the other related variables, thanks to the *arc-consistency on integers* that we used. Arc-consistency technique asserts that these constraints will always be respected for a specific instanciation and try to rule out variable domain values which have no chance to be in a solution. But this technique does not reduce a domain variable to its minimal size ; solutions are complete but they are not all consistent. This is a problem we will have to deal with when generating *topological solutions*.

Figure 4 illustrates a domain reduction propagation with *arc-consistency on integers*. In this example, space *e1* is constrained to be inside a contour of fixed dimensions [0,10]x[0,10]. The domains of length *L* and width *W* are both set to [2,6]. Then, both x1 and y1 domains are automatically reduced to [0,8] and *x2* and *y2* domains are reduced to [2,10]. Let us consider an additional constraint on the surface, *S>12*. A domain reduction of *L* and *W* is immediately achieved, leading to [3,6], because for the value *L=2* (respectively *W*) no consistent value exists in the *W* domain (resp. *L* domain) which respects the constraint: $S = L \times W > 12$.

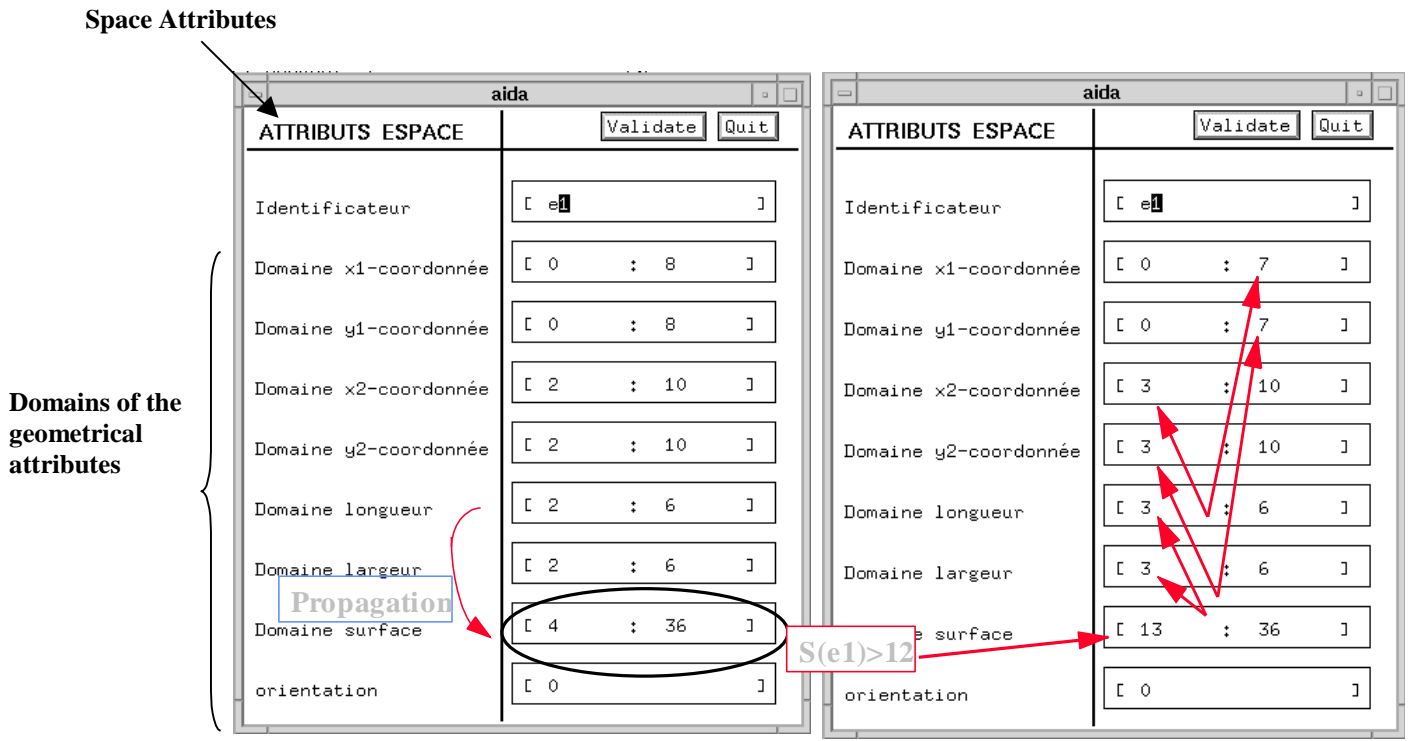

**Figure 4** Illustration of the domain reduction propagation with the arc-consistency technique.

***2.2 Room class* The *room* class defines** spaces other than circulations. It inherits, of course, all *attributes*, *methods* and *class constraints* of the *space* class. This class is characterised by an *orientation* attribute, which is a constrained discrete variable defined over the domain {0°,90°}. Indeed, by stating, for example, that we want one side of the space to measure between 2 and 4, we are not making any particular reference to either the length L, nor the width W. Consequently, it is necessary to consider the two possible configurations (see Figure 5) corresponding to two different orientations. In practice, we pose a choice point leading to two constrained sub-problems. But, for these two branches of reasoning, identical solutions appear during solution enumeration if no special attention is paid to it. For example, if at 0° one of the possible solutions has a length of 3 meters and a width of 2 meters, and at 90° one of the possible solutions has a length of 2 meters and a width of 3 meters, we have a redundant solution (see Figure 6).

The *room* class defines spaces other than circulations. It inherits, of course, all *attributes*, *methods* and *class constraints* of the *space* class. This class is characterised by an *orientation* attribute, which is a constrained discrete variable defined over the domain {0°,90°}. Indeed, by stating, for example, that we want one side of the space to measure between 2 and 4, we are not making any particular reference to either the length L, nor the width W. Consequently, it is necessary to consider the two possible configurations (see Figure 5) corresponding to two different orientations. In practice, we pose a choice point leading to two constrained sub-problems. But, for these two branches of reasoning, identical solutions appear during solution enumeration if no special attention is paid to it. For example, if at 0° one of the possible solutions has a length of 3 meters and a width of 2 meters, and at 90° one of the possible solutions has a length of 2 meters and a width of 3 meters, we have a redundant solution (see Figure 6).

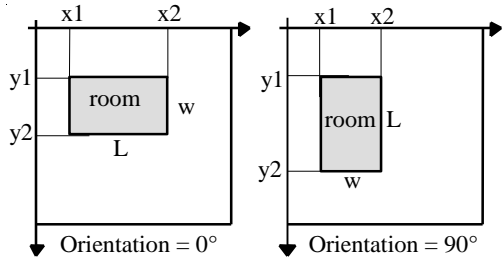

**Figure 5** The two room orientations.

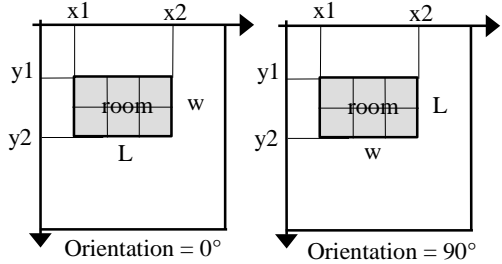

**Figure 6** Example of redundance of one *room*.

To prevent this problem, we have developed a *class constraint* named *orientation redundancy elimination*. This constraint operates as soon as a room *orientation* variable is instanciated, considering the four following cases.

**case 1:** when the *length* and the *width* domains are identical, the *orientation* variable domain is set to {0°}, because the branch corresponding to 90° would only give redundant solutions.

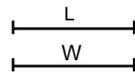

**case 2:** when the *length* and the *width* domains do not overlap each other, then nothing special happens because there is no redundant solution for 90°.

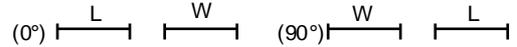

**case 3:** when the *length* and the *width* domains partially overlap each other, then for 90° two sub-problems are considered in order not to enumerate redundant solutions with 0° orientation. They are graphically described by bold lines in the following scheme :

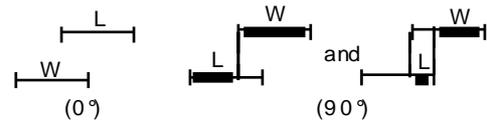

**case 4:** when a domain is completely included in the other one, two sub-problems are again considered for 90°.

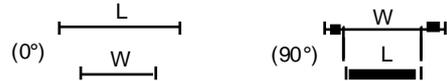

[Charman, 1995] already proposed a constraint of redundance elimination but some redundancies remained.

### 2.3 Staircase class

The *staircase* class is a *space* which is characterised by its orientation. But, contrarily to a *room orientation*, a *staircase* instance has an *orientation* attribute domain of four values {*0°, 90°, 180°, 270°*} (see Figure 7).

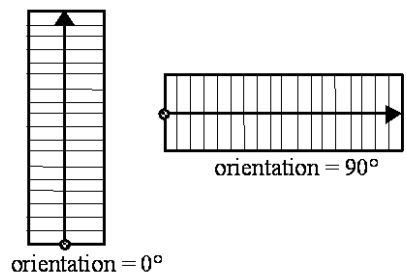

**Figure 7** The four possible orientations of a simple staircase.

The double staircase requires, in addition, an attribute specifying the position of the first step. The initial domain of the constrained variable *position-first-step* is {left, right}. With the four possible orientations, eight constrained sub-problems must be considered (see Figure 8).

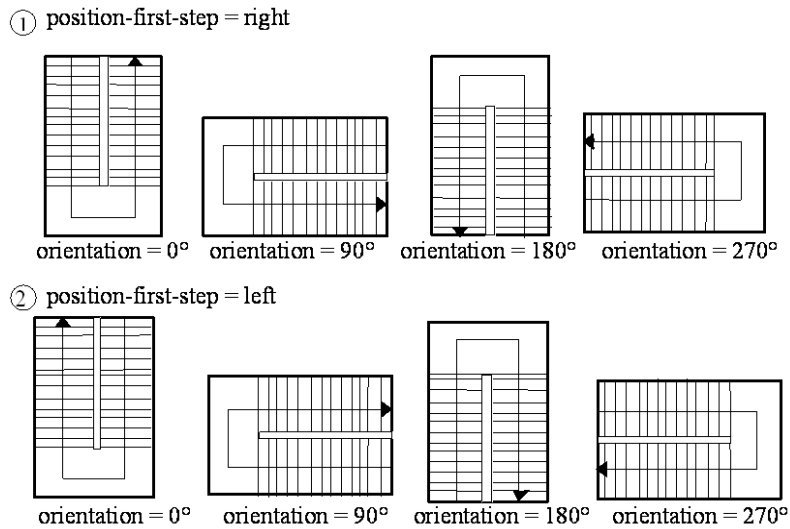

**Figure 8** The eight possible configurations of a double staircase.

## 2.4 Building unit class

A building unit instance is a space instance containing space instances of room, corridor or stairs type. A building unit cannot contain another building unit, then allowing a two-level layout representation. An industrial, sport or scholar complex is composed of building units. But also an apartment building is composed of superimposed building units.

By default, a building unit has no wasted room (between its components) and its components are not superimposed.

## 2.5 Space editor

A simple mouse selection leads to a space creation of floor, room, corridor, simple stairs or double stairs type (see Figure 9).

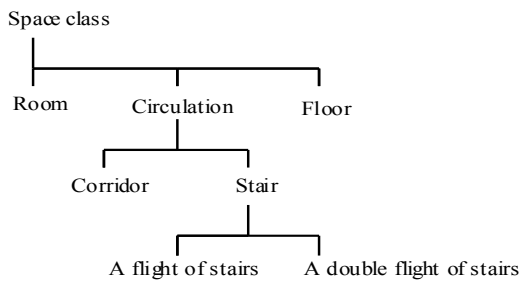

**Figure 9** Hierarchy class in ARCHiPLAN.

Attribute domains are represented as intervals. A *space editor* (already seen in Figure 4) allows to specify interval bounds for all or a part of constrained attributes of a specific space type. As every space is implicitely constrained to be inside the contour of the current floor, a constraint propagation automatically reduces the reference point coordinate domains ($x1, y1, x2, y2$) and the surface area domain $S$. These default values are those of the floor contour when the room is the first to be set inside. As soon as an attribute interval bound value is modified, interval reductions may occur for other attributes according to arc-consistency technique. An attribute interval bound may be modified at any moment after the space creation.

In practice, it is not necessary to define all the variable domains, the $L$ and $W$ domains can suffice.

## 3. ARCHITECTURAL CONSTRAINT MODEL

Our architectural constraint model makes the distinction between *specification constraints* and *implicit constraints* that depend on the fact that:

- they belong to the *functional diagram* and are explicitely declared by the architect (the case of *specification constraints*),

- they are implicitely generated in order to reduce the combinatorial complexity (the case of *implicit constraints*).

## 3.1 Specification constraints

These constraints gather dimensional and topological constraints over *rooms*, *corridors*, *floors, stairs* and *contour*. They are specified by the architects and stored in a *functional diagram*.

Dimensional constraints are applied to the attributes of a single architectural object whereas topological constraints are applied between two or more architectural objects.

### 3.1.1 Dimensional constraints

#### 3.1.1.1 Setting a minimal or maximal domain value

As it has been seen before, setting a minimal or a maximal domain value (especially *width*, *length* and *surface area*) is done through the *space editor*. After the space creation and its initial domain reduction due to its inclusion into the floor contour, additional dimensional constraints on interval bounds are entered if and only if they help to reduce the concerned interval.

*Table 1* presents the dimensional constraints of a *house with two floors* (this benchmark is our own proposition).

**Table 1** Dimensional constraints between spaces (*house with two floors* example). The dimensions are in a module of 0.5 meter (*L-min* stands for *minimum length* and *W-min* for *minimum width*).

| Unit | Area domain values | L-min | W-min | Unit | Area domain values | L-min | W-min |
|---|---|---|---|---|---|---|---|
| Ft_Floor | [320, 320] | 20 | 16 | Sd_Floor | [320, 320] | 20 | 16 |
| Living | [72, 128] | 6 | 6 | Room1 | [48, 60] | 6 | 6 |
| Kitchen | [36, 60] | 5 | 5 | Room2 | [48, 60] | 6 | 6 |
| Toilet/Sh | [16, 36] | 4 | 4 | Room3 | [48, 60] | 6 | 6 |
| Office | [36, 60] | 6 | 6 | Room4 | [48, 72] | 6 | 6 |
| Corridor | [9, 64] | 3 | 3 | Bath1 | [16, 36] | 4 | 4 |
| Staircase | [24, 28] | 4 | 4 | Bath2 | [16, 36] | 4 | 4 |
| Corridor2 | [9, 64] | 3 | 3 | Balcony | [12, 24] | 3 | 3 |

#### 3.1.1.2 Setting a ratio constraint

We developed a ratio constraint between two variables *p1* and *p2*. Practically, it allows to set aesthetic proportions between the dimensions *L* and *W* of a space. But, a ratio constraint may also be set between two length, width or surface area variables of two different spaces. In all cases, this constraint must be considered as a dimensional constraint.

We had to constrain a ratio *p1/p2* to be in a real interval, with constraints on integers. Therefore, we modeled the two interval limits by four positive integers in order to have: $a1/a2 < p1/p2 < a2/b2$. The constraint *ratio(variable#1, variable#2, a1, b1, a2, b2)* leads to two elementary constraints on integers :

$$a1 \times p2 < b1 \times p1$$
$$a2 \times p2 < b2 \times p1$$

For example, if we want to have a *toilets* surface area value between 0.4 and 0.5 times the *shower unit* surface area value, we pose the following constraint : *ratio(toilets.S, shower_unit.S, 2,5,1,2)*.

### 3.1.2 Topological constraints

#### 3.1.2.1 Global overview

As we said, topological constraints allow to specify adjacency, non-adjacency or proximity of a space with another space or with the contour of the current floor. As we will see, the *non-overlapping* between spaces is an implicit constraint systematically considered (even if it can be released) which, consequently, is not considered as a *specification* constraint. The topological constraints can be combined with logical operators such as "OR" and "AND".

In our example of a *house with two floors*, the topological constraints are:

the constraints between floors

- the *first floor* is over the *second fl*oor,
- the *staircase Communicates* between the *first floor* and the *second floor*,

Constraints between spaces of the first floor

- all the spaces of the first floor are *adjacent* to the *corridor* with 1 meter minimum for contact length,
- the *kitchen* and the *living room* are *adjacent* with 1 meter minimum for contact length,
- the *kitchen* is on the *south wall* or on the *north wall* of the *building contour*,
- the *kitchen* and the *Toilet/Shower-unit* are *adjacent*,
- the *living room* is on the *south wall* of the *building contour*,
- all the rooms are naturally lit,
- no space is wasted (the total of the space areas of the first floor correspond to the first floor area),

constraints between spaces of *second floor*

- all the spaces of the *second floor* are *adjacent* to the *corridor* with 1 meter minimum for contact length,
- *room4* and *bath2* are *adjacent* with 1 meter minimum for contact length,
- *room4* and *balcony* are *adjacent* with 1 meter minimum for contact length,
- the *balcony* is *on the south wall* of the *building contour*,
- all the rooms are naturally lit,
- no space is wasted (the total of the *space* areas of the first floor correspond to the *second floor* area),

All these constraints have been introduced into ARCHiPLAN interactively by graph handling and incremental construction (see the specification editor in Figure 17 and the resulting functional diagram in Figure 10).

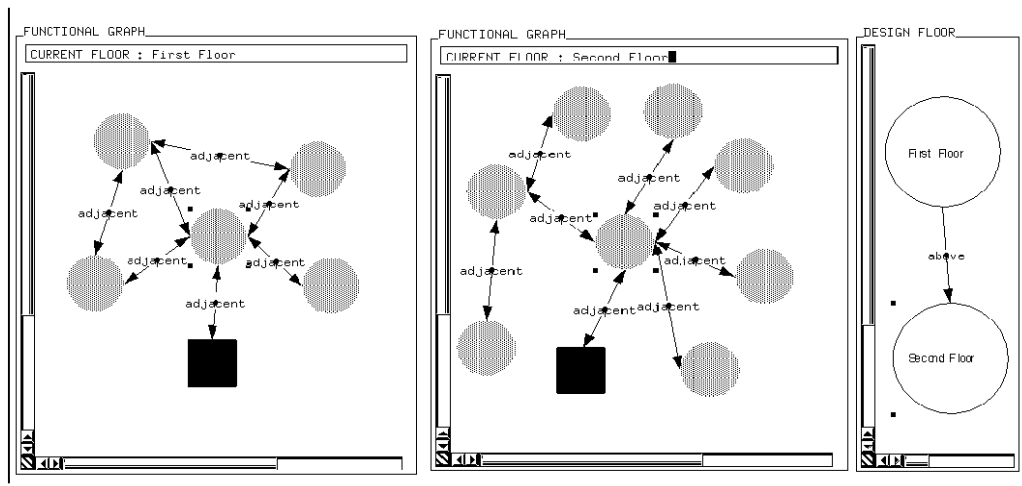

**Figure 10** Functional diagram of the house with two floors example.

### 3.1.2.2 *The generalized adjacency constraint*

The act of conceiving buildings is largely linked with fixing the adjacency between rooms and circulations or fixing the distance between two rooms. All the topological constraints between two spaces (i.e. except constraints between a space and the floor contour) derive from our *generalized adjacency* constraint. Our generalized adjacency constraint is not restricted to direct contact, which is usually called adjacency, but it allows, more generally, to control the relative positioning between two spaces. This constraint is based on two variable notions: the contact length and the distance between spaces.

The contact length *d1* is an integer constrained variable which allows to impose communication between two spaces (see Figure 11). By default, $Min(D(d1))=0$ and $Max(D(d1))=+\infty$, i.e. spaces may have a corner or an entire side in common. In practice, this constrained variable is of course used to impose a minimal communication width for the circulation, which leads to $Min(D(d1))=d1min>0$.

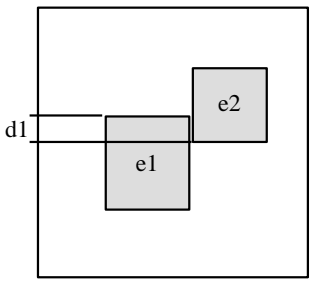

**Figure 11** Contact length *d1* between *e1* and *e2* spaces.

The distance between spaces (*d2*) allows to extend the notion of direct adjacency (contact between two spaces) to a distance specification between two spaces (see Figure 12). This distance *d2* is also an integer constrained variable. By default, its value domain is reduced to the single value 0, and we find the conventional notion of direct adjacency. But, it is often necessary to isolate some storage area (e.g. to store dangerous products) and to impose a safety perimeter; this is expressed as $Min(D(d2))=d2min>0$ and $Max(D(d2))=+\infty$. Generally, we can also impose a maximal distance between two spaces: $Max(D(d2))=d2max>0$ and $Min(D(d2))=0$, or these two constraints can be imposed: $Min(D(d2))=d2min>0$ and $Max(D(d2))=d2max$.

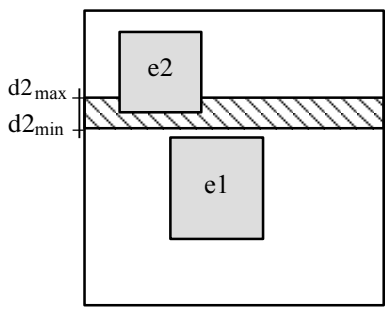

**Figure 12** Distance *d2* between *e1* and *e2* spaces in the case of the constraint *Adjacent to the north*.

Each *adjacent* constraint over a couple of spaces creates a new discrete constrained variable named *adjacency variable* defined over the domain {E, W, N, S}, standing for east, west, north and south. In fact, each adjacent constraint and its consequent adjacency variable poses a choice point (see Figure 13) corresponding to a relative orientation partitioning which is further explained in Figure 19. The adjacent constraint is a "dæmon" constraint for which an instanciation of this adjacency variable, i.e. a relative orientation choice to east, west, north or south, triggers a propagation and consequently a domain reduction thanks to the arc-consistency technique. In chapter 4.1 it will be seen that adjacency variables play a major role in the topological solution enumeration algorithm.

In addition to the general *Adjacent* constraint, four basic adjacency constraints for a specific orientation have been developed : *Adjacent-to-north*, *Adjacent-to-south*, *Adjacent-to-west*, *Adjacent-to-east*. It can be noticed that the *Adjacent* constraint is not simply a disjunction of these four basic constraints because all the solutions corresponding to north-west, south-west, north-east, south-east will be enumerated only once, due to the partitioning (see Figure 13). For the same reason, the following specific disjunctive constraints have been developed: *Adjacent-to-north-west*, *Adjacent-to-south-west*, *Adjacent-to-north-east*, *Adjacent-to-south-east*. They are a mix between a pure disjunction of basic adjacency constraints and the pure partitioning of Figure 13.

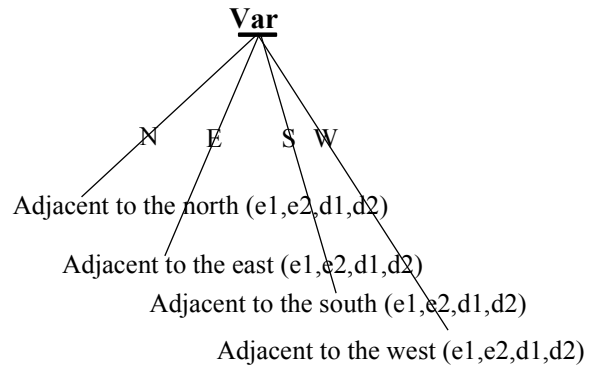

**Figure 13** An *adjacency constraint* creates an *adjacency variable* which is, consequently, a choice point.

### 3.1.2.3 Adjacency with the floor contour

Direct adjacency constraints with the contour of the current floor allow a space to have windows and to benefit from natural lighting. The four basic constraints are: *On-west-contour*, *On-north-contour*, *On-east-contour*, *On-south-contour*. Each constraint simply equals, respectively, the space *x1* or *y1* variable to the *x1* or *y1* contour variable, or a space *x2* or *y2* variable to the *x2* or *y2* contour variable.

We developed special constraints when spaces are constrained to be at contour corners : *On-north-west-contour*, *On-north-east-contour*, *On-south-west-contour*, *On-south-east-contour*. These constraints are roughly considered as disjunctions of two basic direct adjacency constraints with the contour because they again generate an *adjacency variable*, i.e. a choice point with two orientation choices. But, contrarily to a disjunction, the solution at the corners is not enumerated twice.

In the same way, a general *On-contour* constraint can be considered as a disjunction between four orientation choices (see Figure 14). Again, it generates an *adjacency variable* with four possible values {N, S, W, E} but, during enumeration, corner solutions are only enumerated once. This corresponds to the partitioning issue already evoked for adjacency variables.

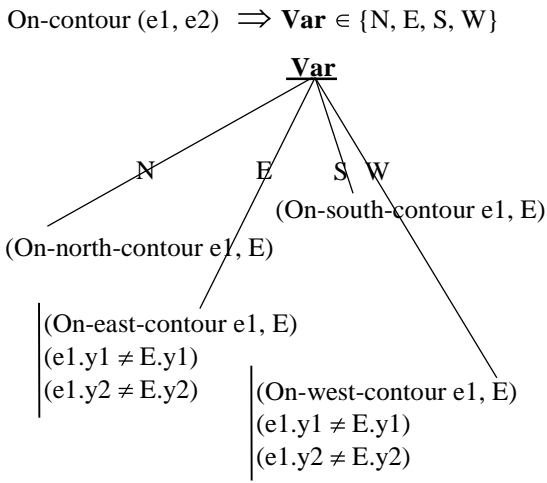

**Figure 14** The *On-contour* constraint poses a choice point.

### 3.1.2.4 Adjacency with staircase

Two specific adjacency constraints between a space and a staircase have been defined. They express that it is forbidden to have a partition in the middle of the first step and in the middle of the last step. Staircase communicate only with one single space, whether for entry or exit. These constraints are triggered as soon as the stairs *orientation* variable is instanciated, an appropriate generalized adjacency constraint (with *d1=0* and a certain *d2*) is then posed (see Figure 15).

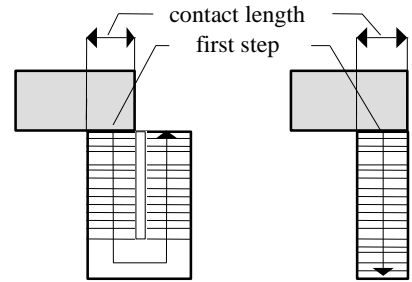

**Figure 15** Adjacency constraint with staircase for climbing.

### 3.1.2.5 Adjacency between corridors

In a primary version of ARCHiPLAN, a corridor is also a rectangular space. This is rather restrictive, as architects do not a priori know the shape of the corridor. Therefore, an architect must actually successively study a one-rectangle and a two-rectangle corridor issue. The consistency of a two-rectangle corridor, composed by two elementary adjacent corridors, is given by a special adjacency constraint. This constraint avoids solutions that are geometrically identical (see Figure 16). Such problems may occur when two corridors form in fact a single corridor. Each time that two corridors *c1* and *c2* of equal width are aligned, the length of *c1* is set to the minimal value of its domain as long as the length of *c2* is inferior to the maximal value of its variable domain. Corridors can be a maximum of two rectangles (2-rectangle corridors), not because of the "isothetic rectangle" but because of the constraint composition explained in section 3.1.3.

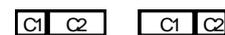

**Figure 16** Example of geometrically identical solutions for two corridors.

### 3.1.3 Specification constraints composition

A conjunction of constraints does not present any problem in constraint programming. Actually, the more constraints there are the more efficient the resolution is. Contrarily, a disjunction enlarges the search space, so one must take care to only enlarge it at its minimum. In ARCHIPLAN, the following disjunctive form is supported:

(*e1* (*constraint-conjunction#1*) *e2*) OR
(*e1* (*constraint-conjunction#2*) *e3*) with $e2 \neq e3$ [2]

where e1, e2 and e3 are three spaces and OR is inclusive. For example, one can have :
*(Corridor Adjacent-to-west kitchen) OR (Corridor Adjacent-to-east Living-room)*

This disjunction always creates a *choice point* with two reasoning branches. But in the case where " *e1 is Adjacent to e2 OR e3* ", the solutions for which *e1* is adjacent to both *e2* and *e3* are enumerated twice (one for each reasoning branch). We implemented a constraint whose principle is, in the second reasoning branch, to only enumerate solutions which have not yet been enumerated in the first reasoning branch. We did not generalize this disjunction constraint to complete propositional formulas (combinations of AND, OR, NOT) because of the complexity of eliminating redundant solutions.

Let us recall that, in all constraints implementation, we were always concerned with the fact that a solution should be enumerated only once (so that for our future *topological solutions* be distinct, see chapter 4). Moreover, this simple disjunction is important because it allows adjacency with a two-rectangle corridor (of L-shape or T-shape), a space being adjacent to one part of the corridor at least.

### 3.1.4 Specification editor

Specifications of an architectural problem are represented into a graph, called *functional diagram* by the architects. Graph nodes stand for spaces and support *dimensional constraints* (except *ratio constraint*) over them. Graph links support *topological constraints* (between two or more spaces) and *ratio constraints* between two constrained attributes of the same space (reflexive link) or of two different spaces.

The general *specification editor* allows to build, cut, paste, save, load and graphically edit this *functional diagram*.

We already mentioned that *dimensional constraints*, except *ratio constraint*, were defined with the *space editor*. The *space editor* is activated each time the mouse clicks on a graph node.

The general *specification editor* (see Figure 17) is split into three main panels :

- the building unit layout window (to the right),
- the current building unit window (in the middle),
- the topological constraints specification panel (to the left).

---

2 The case where e2=e3 is already considered by the specific evolved generalized adjacency constraints: Adjacent-to-north-west, Adjacent-to-south-west, Adjacent-to-north-east, Adjacent-to-south-east and Adjacent.

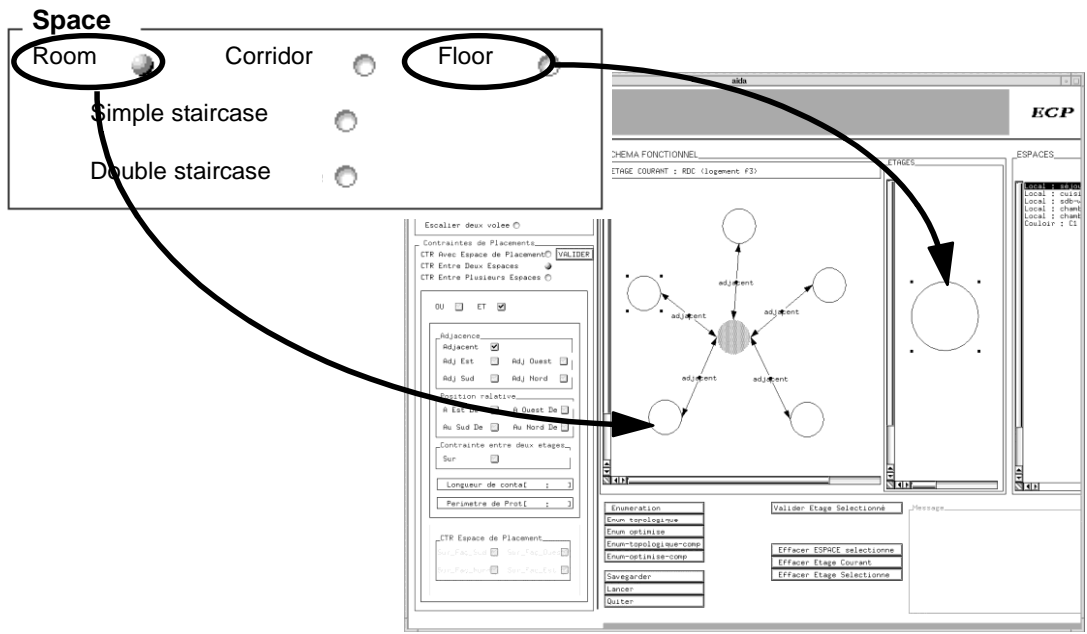

**Figure 17** The specification editor of ARCHIPLAN.

The *building unit layout window* (to the right) (see Figure 17) is dedicated to the definition and specifications between *building units* of the same floor and/or different floors. Building units may be constrained by generalized adjacency constraints as well as rooms between them inside a particular building unit. But, when two building units are consecutive floors of an apartment building, two special constraints are allowed:

- building units are constrained to have a similar contour ; this is the *superimposition* constraint,
- a *stairs constraint* can link two consecutive floors, i.e. the stairs have the same size on both floors but in taking into account the position of the first and the last step for an *adjacency* constraint (see further) with a corridor or a room. This constraint appears as a graph link between building units and staircase instances are automatically created in each building unit.

In a first stage, building units and stairs are created and constrained in the right-hand window. In a second stage, and for each building unit, rooms and corridors are created and constrained in the *current building unit window* (in the middle, see Figure 17). Spaces are created inside the current building unit. One can switch from a current building unit to another by a simple selection on the appropriate node in the *building unit layout window*. By default, the following *implicit constraints* are automatically posed : *inclusion in the contour*, *non-overlapping* between spaces, *contour total recovery* (see next chapter) unless the architects intentionally release them. For example, the *contour total recovery* constraint may be released, solutions with extra-space are then proposed. In such a case, the criterion of corridor surface area minimization (see chapter 5.3) is extended to this extra-space area.

*Topological constraints* are edited from the left-hand panel of the *specification editor* (see Figure 17). This zone concerns the main constraints of a space in relation with the contour (*On-west-contour*, *On-north-contour*, *On-east-contour*, *On-south-contour*, *On-contour*) and the main *generalized adjacency constraints* between two spaces (*Adjacent-to-north*, *Adjacent-to-south*, *Adjacent-to-west*, *Adjacent-to-east*, *Adjacent*). In Figure 18, we see that the constraints composition can simply be specified by quick interactions on some buttons. We saw that, in the general case where (*e1* (*constraint-conjunction#1*) *e2*) OR (*e1* (*constraint-conjunction#2*) *e3*) with *e2≠e3*, a special constraint was activated for posing a choice point without enumerating redundant solutions. When *e2=e3* a constraint analyser detects such a case and activates the appropriate constraint among : *Adjacent-to-north-west*, *Adjacent-to-south-west*, *Adjacent-to-north-east*, *Adjacent-to-south-east* and *On-north-west-contour*, *On-north-east-contour*, *On-south-west-contour*, *On-south-east-contour*.

For example, it is possible to state that *toilets* are *Adjacent* (directly, i.e. *distance between spaces d2* is instancianted to 0) to *kitchen* with a *contact length l1* OR to *bathroom* with *contact length l2*[3]. In practice, the notions of *contact length* and *distance between spaces* proved to be very flexible and powerful. For example, the *non-adjacency* constraint is defined by setting the minimal value of *d2* to 1. By default, the editor proposes the following bounds for d1 and d2 intervals : *Min(D(d1))=0, Max(D(d1))=+∞* and *Min(D(d2))=0, Max(D(d2))=0*. These values correspond to a conventional direct adjacency without any particular minimal *contact length* value.

---

3  With a distance increment of 0.25, one can think of a minimal contact length value of 4 corresponding to a communication of 1 meter.

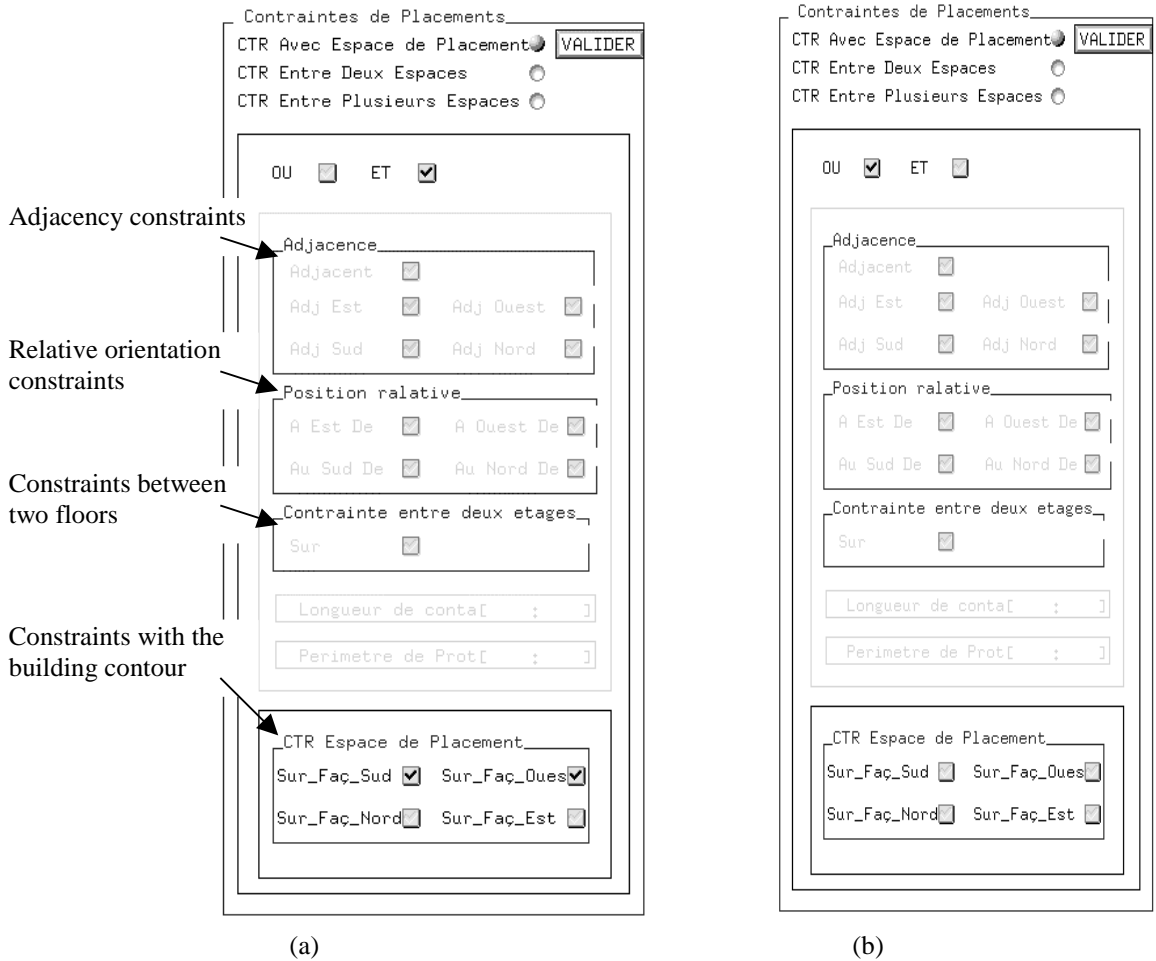

**Figure 18** Selection zone for the topological specifications definition in ARCHiPLAN.
Case (a) represents the constraint *On-west-contour* AND *On-south-contour*.
Case (b) represents the constraint *On-west-contour* OR *On-south-contour*

As soon as a constraint is defined, it appears as a graph link or several graph links if the constraint is a disjunction. All these constraints may be modified by a simple click with the middle button of the mouse (the left-hand button is for a drag and drop of the graphic item).

All these constraints have been introduced in ARCHiPLAN interactively by graph handling and incremental construction. This functional diagram of the example of the house with two floors example has already been given in Figure 10.

### 3.2 Implicit constraints

#### 3.2.1 Default constraints

These constraints are considered by default but they can be released for special cases.

#### 3.2.1.1 Inclusion in the current building unit contour

This simple constraint consists of four conjunctive inequalities over $x1, x2, y1, y2$ in order to be inside of the current building unit contour.

#### 3.2.1.2 The contour total recovery constraint

The *contour total recovery constraint* expresses the fact that there is no lost space in a building unit and therefore, that the sum of space surface areas of a given building unit equals the whole building unit surface area.

#### 3.2.1.3 Non-overlapping constraints

A *non-overlapping constraint* expresses the fact that a space cannot overlap another space; it is automatically posed between all pairs of rooms. Of course, pairs of rooms which are already constrained to be directly adjacent verify the *non-* *overlapping constraint*. Figure 19 shows the position permitted for *e2.x2* and *e2.y2*[4] by the *non-overlapping* constraint between spaces *e1* and *e2*. This constraint is dependent on the *minimal space dimension* notion. The minimal space dimension is, at any moment, equal to the smallest dimension value (width or length) of all spaces. This value is used in order to constrain two spaces to be adjacent, or to be separated by a sufficient distance which allows another space to be inserted in between. As the *Adjacency* constraint, the *non-overlapping* constraint introduces a new *non-overlapping variable* with four values {$E,W,N,S$}. This variable divides the space surroundings into four parts (see Figure 19) but not symmetrically. Indeed, we observe that the N and S values give more solutions that the E and W values. It is the instanciation of these *non-overlapping variables* and the *adjacency variables* which, if proven consistent, gives a *topological solution*. We can consider the following equivalence:

*non-overlapping (e1,e2)=Adjacent (e1 e2 d1 d2)* (1) with $d1 \in [0+\infty]$ and $d2 \in [0 +\infty]$.

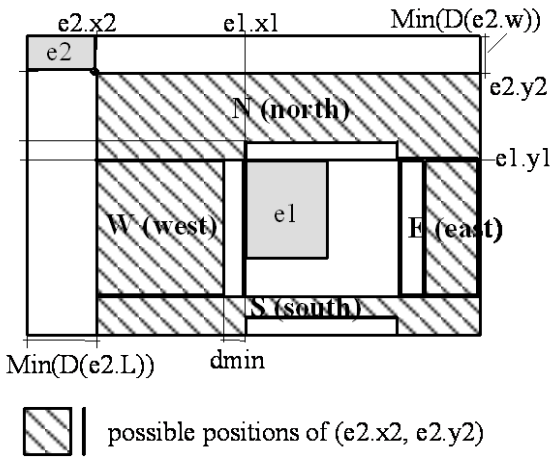

**Figure 19** Positions permitted for point (*x2, y2*) of space *e2* after the non-overlapping constraint applied between space *e1* and *e2*. The partitioning of the surroundings of a space in {E,W,N,S} is given.

### 3.2.2 Research space reduction constraints

These constraints which allow a drastic reduction of the combinatorial, are specific to our approach. They regroup:
• the incoherent space elimination constraint,
• the symmetry constraints,
• the topological reduction constraint,

---
[4] *e2.y2* represents the constrained variable *y2* of space *e2*.

• the orientation propagation constraint.

#### 3.2.2.1 The incoherent space elimination constraint

This constraint is also dependent on the *minimal space dimension* notion. The aim is to constrain each space to be either directly adjacent to the building unit contour or to be distant from a certain value, for another space to be inserted in between. This value is distance *dmin=Min(Lmin,Wmin)*. This constraint is activated if and only if the *contour total recovery constraint* is activated too. Figure 20 shows the positions permitted for *(x2,y2)* point of space *e1* with this constraint, relatively to the building unit contour. The constraint algorithm is described in Figure 21.

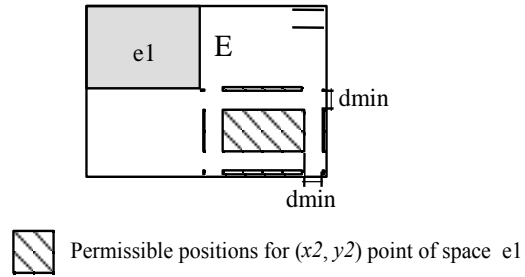

**Figure 20** Permissible positions for (*x2,y2*) point of space e1 for the incoherent space elimination constraint.

```
Constraint Eliminate-inconsistency (IN: e1, E)
For i varying from 1 to (dmin - 1)
    →      e1.x1 ≠ E.x1+i
           e1.x1 ≠ E.x2 - e1.L + i
For j varying from 1 to (dmin - 1)
    →      e1.y1 ≠ E.y1+j
           e1.y1 ≠ E.y2 - e1.W + j
End Constraint
```

**Figure 21** The *incoherent space elimination constraint* algorithm. E is the building unit space.

#### 3.2.2.2 The symmetry constraints

The *symmetry constraints* are meant to avoid functionally identical solutions by solution combinations over spaces of the same type and with the same constraints: same initial domains and same topological constraints with other spaces. For example, let us take a house with three similar rooms (for children) having the same initial dimensional domains and the same direct adjacency constraint with the corridor.

In order to rule out symmetrical combinations between two spaces *e1* and *e2*, it is sufficient to constrain *e1.x1* to always be lower than or equal to *e2.x1* and when *e1.x1=e2.x1*, one must impose *e1.y1<e2.y1* (see Figure 22). This procedure is applied for n symmetrical spaces, the algorithm is described in Figure 23.

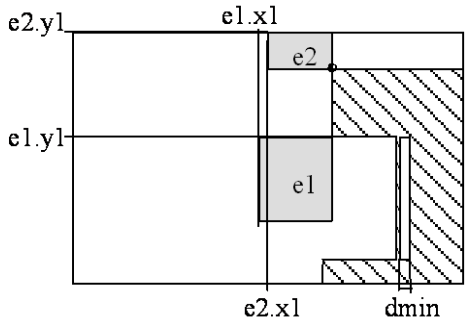

**Figure 22** Permissible positions for *(x2,y2)* point of space *e2* for the *symmetry* and *non-overlapping* constraints.

---

**Symmetry Constraint** (*I*: List-of-symmetrical-spaces)
$n$ = length (*l*)
**For** i varying from 1 to *n*
  $e_i$ = element-of (i, *l*)
  **For** j varying from (i + 1) to *n*
    $e_j$ = element-of(j, *l*)
    → $e_i.x1 \le e_j.x1$
    **When** $V(e_i.x1) \equiv V(e_j.x1)$
      → $e_i.y1 < e_j.y1$
**End Constraint**

---

**Figure 23** The *symmetry constraint* algorithm. E is the building unit space.

This above elementary *symmetry constraint* algorithm has been generalized to the different orientations of a room, the *orientation* attribute having {0°,90°} initial domain. In the case where two symmetrical rooms have two possible orientations, the previous elementary symmetry constraint is triggered each time both orientation attribute values are equal. When (*V(orientation.e1)=0°* and *V(orientation.e2)=90°*), there is no symmetrical solutions. But all solutions corresponding to *V(orientation.e1)=90°* and *V(orientation.e2)=0°* have been enumerated in the previous case *V(orientation.e1)=0°* and *V(orientation.e2)=90°*. In order to rule out these redundant solutions, we will consider only the case when the orientation attribute values are different only once. Figure 24 illustrates the symmetry constraint generalized to different orientations.

---

**GenSymmetry Contrainte** (*I*: List-of-symmetrical-spaces)
$n$ = length (*l*)
**For** i varying from 1 to *n*
  $e_i$ = element of (i, *l*)
  **For** j varying from (i + 1) to *n*
    **When** $V(e_i.orientation) \equiv V(e_j.orientation)$
      (**Symmetry** (list(ei, ej)))
    **When** $V(e_i.orientation) \neq V(e_j.orientation)$
      **When** $V(e_i.orientation) \equiv 90°$
        → $V(e_j.orientation) \neq 0°$
**End Constraint**

---

**Figure 24** The symmetry constraint generalized to different orientations.

### 3.2.2.3 *The topological reduction constraint*

The *topological reduction constraint* operate when adjacency constraints with the building unit contour exist. The principle is : when a space *e1* is *On-north-contour*, no other space can be to the north of space *e1*. The topological reduction constraint rules out the {*N*} value of the domains of the *(n-1) non-overlapping variables* relative to space *e1* (coming from the *non-overlapping constraints*). When reducing these variable domains, we directly eliminate some inconsistant topologies.

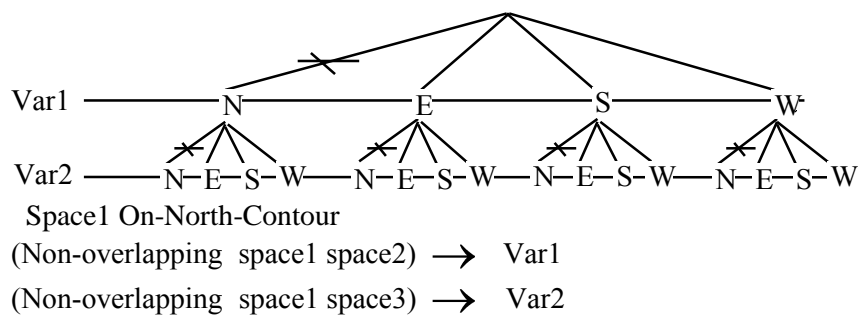

Space1 On-North-Contour
(Non-overlapping space1 space2) → Var1
(Non-overlapping space1 space3) → Var2

**Figure 25** Elimination of the {*N*} value from the *non-overlapping variable* domains.

This *topological reduction constraint* is similar for the three other orientations :
- if a space is *On-South-contour*, no space can be to the South of it,
- if a space is *On-West-contour*, no space can be to the West of it,
- if a space is *On-East-contour*, no space can be to the East of it.

### 3.2.2.4 The orientation propagation constraint

The *orientation propagation constraint* uses the orientation transitivity property to automatically instantiate *non-overlapping variables*. For instance, if space e1 is North-of e2 and e2 is North-of e3, thus e1 is North-of e3. We developed such a transitivity constraint for relative orientations of North and South. We did not develop equivalent constraints for East and West because the non symmetrical partitioning into {N, E, O, S} does not guarantee the transitivity (see Figure 19). This partitioning considers north-west and north-east as a part of North, and south-west and south-east as a part of South (e.g. if space e1 is East-of e2 and e2 is East-of e3, e1 can be North or South-of e3).

## 4. THE TOPOLOGICAL SOLUTION LEVEL

### 4.1 The topological solution definition

We wanted our *topological solution* definition to correspond to the architect's notion of sketching where the adjacency between spaces is defined but where space sizes are imprecise. This geometrical precision is treated in the next chapter with geometrical solutions where all space attributes are instanciated.
Finally, we converged to the following definition of a topological solution:

> Each space layout Constraint Satisfaction Problem (CSP) where the n.(n-1)/2 (n being the number of spaces) non-overlapping variables and adjacency variables are instanciated and which remains geometrically consistent (i.e. for which at least one geometrical solution exists) is a topological solution.

At this stage, value domains have undergone reduction but dimensional variable domains are not necessarily reduced to a unique value (i.e. instanciated), only non-overlapping and adjacency variables have been instanciated. We therefore believe that there can exist several geometrical solutions consistent with this topological solution. What is important to say is that the whole constraint model has been developed in such a way that a geometrical solution can derive from only one topological solution. This is particularly the case for :
- the *non-overlapping* and *generalized adjacency constraints* and their non-symmetrical relative orientation partitioning,
- the *orientation redundancy elimination class constraint*,
- the *symmetry constraint*.

Consequently, topological solutions are distinct equivalence classes of geometrical solutions. This is an important property which allows to make design decisions over topological solutions (elimination or further study of complete solutions classes). If it is impossible to satisfy the topological constraints the user can modify the module of the instanciation (if the module was 10cm it could be possible to satisfy the topological constraints for a module equal to 5cm or 1cm).

The verification of a topological solution consistency amounts to the research of a first geometrical solution. This research uses the same algorithm as the geometrical solution optimization algorithm presented in a further chapter.

### 4.2 The two topological enumeration heuristics

The topological solutions enumeration algorithm is based upon two enumeration heuristics which were detailed in a previous paper (see [Medjdoub & Yannou] for more details).

Traditionally, the constraint programming approaches that have been developed enumerate the geometrical solutions straightforwardly. The image that can be used is successively to dimension and to place each space in the building unit contour which is initially empty. The heuristics that have been already proposed corresponds essentially to the choice of the next space to dimension and to place (Andre, 1986; Eastman, 1973; Pfefferkorn, 1975), and sometimes to the choice of the location where the first space considered must be placed (Charman, 1994).

Our approach is different because it enumerates the topology in a first instance. This does not correspond to the dimensional space attributes instanciations but to instanciations of *non-overlapping* and *adjacency variables* relatively to already placed spaces.

A first heuristics consists in choosing the next space to deal with. It is based on the choice of the currently most constrained space with the building unit contour and with the already placed spaces.

A second heuristics consists in choosing the variable instanciation order (among *non-overlapping* and *adjacency variables*).

Both heuristics have been generalized to several building units.

### 4.3 The topological graphical representation

Naturally, we tried to represent the topological solutions graphically by adopting average values of the value domain of the space attributes (*x1, y1, x2, y2*). We then noticed the striking resemblance between such graphic representations and sketches that are made by hand by architects in preliminary design. Similarly to a sketch, the graphic representation of a topological solution reveals a slight overlapping of rectangles (see Figure 26). In the example of the *house with two floors*, 49 topological solutions are found and are displayed by ARCHiPLAN (see Figure 26).

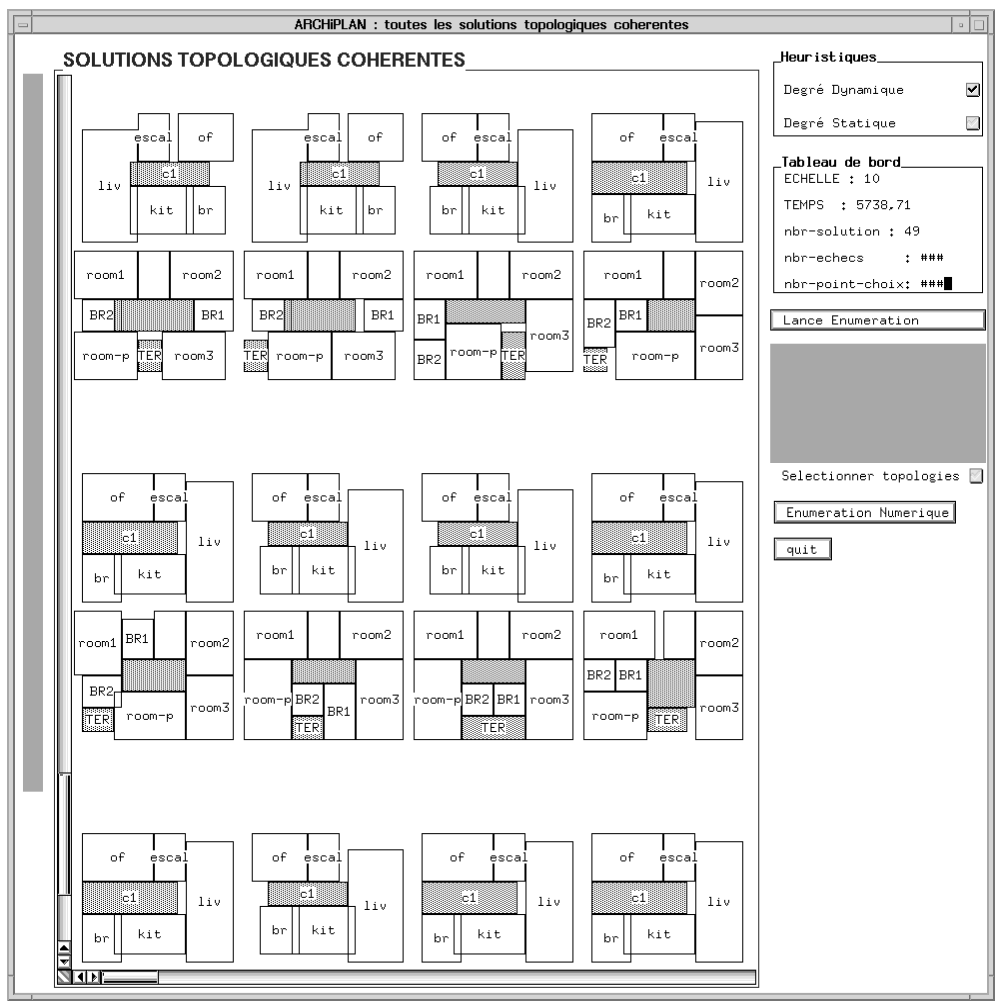

**Figure 26** Some topological solutions of the *house with two floors example* among the 49.

## 4.4 The topological solutions manager

ARCHIPLAN is an automatic conceptual solution generator. It can be functionally compared to the ABD approach [Schwarz et al, 1994]. But both approaches are not identical (see chapter 7.2). An architect makes efforts to imagine some topological sketches from the functional diagram, but whereas architects are creative and innovative, a computer is exhaustive and runs fast. Such a conceptual solutions generator lets the architects embrace the "fields of possibles" in a glance. Far from imposing a specific design, such an approach immediately shows them what is not possible. For example, although architects can think they have found a correct sketch (or topological solution) because topological orientations are checked, it can reveal itself as an incoherent solution when taking geometrical constraints into account. In ARCHIPLAN, only consistent solutions are presented. In the same way, ARCHIPLAN provides some interesting functionalities for the conceptual design stage:

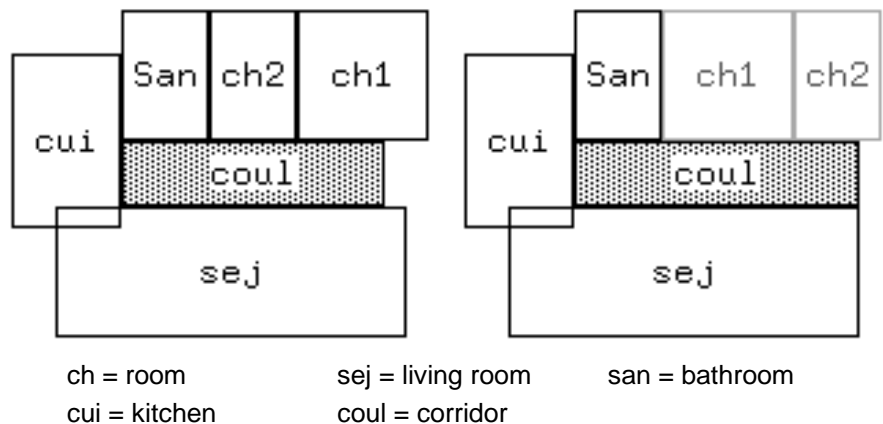

ch = room     sej = living room     san = bathroom
cui = kitchen     coul = corridor

**Figure 27** A color standard stresses topological differences between a solution and the previous one in the enumeration.

- **Comparing two topological solutions.** ARCHIPLAN stresses the topological differences between, by default, two consecutive topological solutions in the enumeration or between two selected solutions, thanks to a color standard (see Figure 27).
- **Sorting topological solutions from several criteria**. For example, the architects ask to sort the only topological solutions which will permit to have the surface area of a space lower than 20 m$^2$. ARCHIPLAN allows this type of hypothetical reasoning. The $S<20$ constraint is added to all topological solutions and constraint propagation and consistency checking (in finding a first geometrical solution) are carried out. The sorted solutions are those which lead to consistent solutions after applying the additional constraint. Finally, initial constrained systems for sorted solutions are restored after this hypothetical reasoning.

The aim of these two first functionalities is to help the architects to choose among the conceptual solutions in order to study the most interesting solutions in more details.
- **Better apprehending a topological solution.** The architects can benefit from capital information when editing space attributes of a topological solution. The narrower the domains are, the more constrained the system is, and the less additional constraints have a chance to be accepted.
- **Numerically exploring a topological solution**. ARCHIPLAN lets the architects manually explore the numerical space of a topological solution. In the case where no explicit cost function exists but where expertise is in the architect's mind, the architects can test hypotheses successively and come backwards at any moment. It proceeds by domain reductions or instanciations, the system carrying out constraint propagation and geometrical consistency checking at every step. This incremental design by successive refinements is allowed by the constraint programming facilities. This functionality must be compared with the variational geometry for which we have no information about the remaining degrees of freedom of the geometry. Moreover, it can be decided, at any moment, to enumerate all the remaining geometrical solutions (with the risk of leading to a combinatorial explosion). Geometrical solutions are then collected into a geometrical solutions manager detailed further.

## 4.5 Checking of consistency

After our definition, a topological solution must be proved geometrically consistent. At least one geometrical solution must be enumerated. We wanted to measure the specific efficiency of the arc-consistency technique. For that purpose, we compared the number $N1$ of potential topological solutions after constraint propagation to the number $N2$ of effective topological solutions after the geometrical consistency checking. The relevance of arc-consistency on this issue is given by the ratio of $N2$ over $N1$, i.e. the percentage of effective solutions in potential solutions.

We adopted the following problems :

- The *Pfefferkorn* problem (Pfk) [Pfefferkorn,1975]: Six rectangles of fixed dimensions : 6x2, 4x2, 2x3, 2x3, 2x3 et 2x1 must be assembled into another rectangle of fixed dimensions 8x5. The rectangles have a unique 0° orientation.
- The *Laurière* problem (Lr) [Laurière, 1976]: It is a variant of the *Pfefferkorn* problem. Here, rectangles can have two orientations.
- The *Tong* problem (Tng) [Tong, 1987]: Four rectangles where all the sides vary from 4 to 9 and must be placed into a 9x9 rectangle.
- The 9 perfect squares (Col9) of *Charman* [Charman, 1995] inspired by [Colmerauer, 1990].
- The *Maculet* problem (Mac) [Maculet, 1991]: this is a one-family dwelling problem with ten spaces in a fied building unit contour. Detailed constraints are given in the *case studies* chapter. It was the most complex problem.

**Table 2** Relevance of the arc-consistency technique for a space layout planning problem.

Problem  Tng  Pfk  Lr  Col9  Mac

| Problem | Tng | Pfk | Lr | Col9 | Mac |
|---|---|---|---|---|---|
| Number of rooms | 4 | 6 | 6 | 9 | 10 |
| N1 potential | **24** | **24** | **72** | **4** | **345** |
| N2 consistent | 4 | 24 | 72 | 4 | 72 |
| N2/N1 | 17 | **100** | **100** | **100** | 21 |
| T1 : all potential solutions | **0,65** | **11,1** | **72** | **4,56** | **3245** |
| T2 : first geometrical solution | 0,085 | 0,09 | 0,09 | 0,09 | 0,20 |
| T2' : best geometrical solution | 0,103 | 0,111 | 0,12 | 0,11 | 0,205 |

For example with initially instanciated spaces, topological solutions are obviously consistent because they are also geometrical solutions. For the two remaining problems, a ratio of about 20% reveals a rather good efficiency for arc-consistency. It had been noted that time $T2$ for finding the first geometrical solution is slightly lower than time $T2'$ for finding the best geometrical solution (see next chapter for optimization algorithm and cost functions). We will see further why this surprising property exists for the native " branch and bound " optimization algorithm of a constraint programming package. In consequence, we decided to directly adopt the optimal solution search for the geometrical consistency checking rather than the first solution search.

# 5. THE GEOMETRICAL SOLUTIONS LEVEL

## 5.1 The detailed design

For the architects each topological solution is a space layout planning principle. ARCHIPLAN enumerates all these conceptual solutions starting from a functional diagram. As they are generic classes of geometrical solutions, their number is not so important even for problems of practical size (20 rooms) and for low-constrained problems. This number typically varies from ten to one hundred, a number easily apprehendable as a whole. Three possibilities exist to tackle with the detailed design process:

- For a low number of attractive topological solutions, the architects can try to refine them by adding subjective constraints that were not initially in the functional diagram in order to converge step by step towards an instanciated solution that we call a geometrical solution. We saw previously that ARCHIPLAN lets the architects lead this incremental design approach with sorting and hypothetical reasoning mechanisms.
- A second approach is to let the architects express a cost function to find the best geometrical solution corresponding to each topological solution of interest. The limit of this approach is to be able to explain an exhaustive set of criteria and to be able to weigh their respective importance. At present, ARCHIPLAN proposes to minimize the total length of walls and the surface area of corridors. In the short term, we aim to develop grid criteria, noise minimization criteria, cost criteria, flow lengths minimization criteria and insulation maximization criteria.
- The third solution is to straightforwardly enumerate all the geometrical solutions. The risk remains the high number of solutions and the very long reply time. The time to find all the solutions of a constrained problem is greater than that required to find the best solution with the "branch and bound" algorithm already evoked

## 5.2 Optimization algorithm

Very few optimization approaches in architecture exist, let us mention Ligett's [Ligett, 1991]. Our optimization approach consists in minimizing an objective function, called *cost function*, composed as a weighted sum of criteria. Our " branch and bound " optimization method leads to the determination of the global optimum (eventually global optima) of a topological solution. This is not the case of expert systems approaches or evolutionary approaches (Damski and Gero, 1997; Jo and Gero, 1997) which only lead to "satisfactory solutions".

The *" Branch and Bound " algorithm* is based on the *enumeration algorithm* which builds a depth-first research tree.

For the *enumeration algorithm*, each choice point in the research tree corresponds to a variable choice (for example $x$) among those which have not been instanciated yet. Each branch corresponds to a particular instanciated value (for example $v$) in the variable domain. Coming down the tree consists in adding the constraint $x = v$, coming up or *backtracking* consists in releasing this constraint, i.e. in restoring the ancient constraint set. Each addition of a constraint triggers a constraint propagation which reduces the domains of the remaining variables to instanciate. When a domain becomes empty, no solution exists in this branch and a backtrack is carried out. In the enumeration algorithm the order of the choice of variables considerably influences the size of the tree and consequently the overall duration of the enumeration process. The algorithm leading to the choice of variables is called the *variable choice heuristics*. Typically it consists in choosing first the most constrained variable in order to quickly backtrack if no solution exists. This heuristics is dynamic, i.e it is not applied only once leading to a fixed global variable ordering but it is applied at each step for the remaining set of non-instanciated variables. The *heuristics* term is somewhat confusing because this enumeration algorithm provides the complete solution set ; there is no approximation. The second *value choice heuristics* does not influence the overall enumeration process duration at all.

With the previous enumeration algorithm, the *" branch and bound " algorithm* consists in finding, a first solution S1. Let us recall that the objective is to find the solution with the lowest cost function value. This is why the new constraint *Cost-function<Cost-function(S1)* is applied, and this constraint is not released when backtracking. This new constraint provokes domain reductions. Better the solution S1 is (i.e. *Cost-function(S1)* is low), more efficiently the domains reduction are. A second solution S2, better than S1, can be found and a stronger constraint is posed : *Cost-function<Cost-function(S2)*, and so on until all the values have been tested. We can conceive here that the optimization process duration is related to the ability to find a correct solution immediately and thus to the *value choice heuristics*.

For the choice of spaces, we have developed a *variable choice heuristics* and a *value choice heuristics*. The *value choice heuristics* is based on a first building unit choice heuristics.

For the issue of the *consistency checking*, the fact that the optimization process duration and that the first solution search process duration are very close is due to two reasons:

- we have a satisfactory *value choice heuristics*,
- the actual optimization criteria (see further) of the cost function are linear criteria of space variables. The first solution S1 provokes already large domain reductions even if S1 is not so satisfactory.

The fact that both durations are small is also due to two reasons:

- a topological solution is already a very constrained problem for which variable domains have strongly been reduced,
- the *inconsistent space elimination constraint*, which is a dynamic constraint, efficiently prune the research tree. Indeed, as soon as a space is instanciated, if the minimal distance to the building unit contour is lower than the lowest side of the remaining spaces to be placed, it provokes a backtrack.

### 5.3 Optimization criteria

Usually, an architect wants to favor room surface areas rather than corridor surface areas. For that purpose, we developed the *corridors' surface area criterion* given by the following formula :

$$C\_circulation = \sum_{i=1}^{n} Circulation_i.S$$

But this minimization can also take spaces other than corridors into account.

In order to minimize the amount brickwork, the *total length of walls* must be minimized, comprising internal partitions and external walls (building unit contours are not necessarily initially instanciated).

The term : $2(e_i.L + e_i.W)$ represents the sum of all internal space perimeters. Hence, this term equals the sum of the length of external walls and twice the length of internal partitions because a partition length is considered twice in two perimeters. Thus the *internal partitions' length criterion* is given by the following formula:

$$(e_i.L + e_i.W) - E.L/2 - E.W/2$$

The *internal walls' length criterion* and the *external walls' length criterion* must be weighed by their respective linear costs, leading to the formulas :

$$C\_L\_Internalwall = \left[\sum_{i=1}^{n}(e_i.L + e_i.W) - \frac{E.L + E.W}{2}\right] \times Cost_{internal\ wall}$$

$$C\_L\_Externalwall = (E.L + E.W) \times Cost_{external\ wall}$$

$$C\_Lwall = C\_L\_Externalwall + C\_L\_Internalwall$$

When these costs are not a priori known, an solution acceptable for both internal partitions and external wall can be :

$$(e_i.L + e_i.W) + E.L/2 + E.W/2$$

Numerous other criteria should be introduced to enrich the decision making process. A great flexibility for such extensions is that the geometrical optimization algorithm is not modified at all thanks to constraint programming techniques which well separate constraints from enumeration and optimization algorithms.

### 5.4 The cost function editor

From a viewpoint of CAD user-friendliness, it is very simple to offer the architects an interactive tool to compose his cost function by tuning the relative importance of the evoked elementary criteria (see Figure 28). As has been mentioned, it is very easy to extend and customize the criteria library.

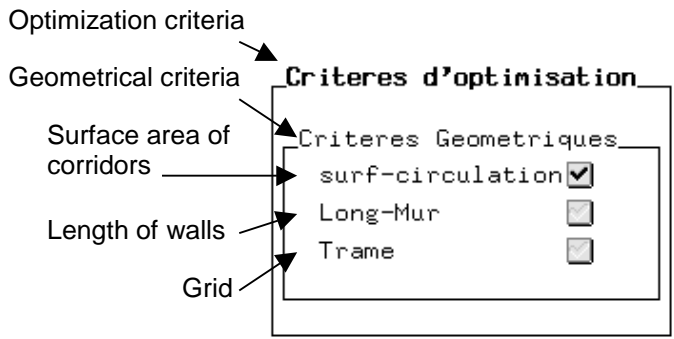

**Figure 28** Choosing to minimize the *total corridor surface area* criterion.

## 5.5 The geometrical solutions manager

Most of the time, the optimization of a topological solution leads to one (more seldom several) optimal geometrical solution. The *n* (or more) optimal geometrical solutions corresponding to the *n* topological solutions are globally displayed in a *manager of best geometrical solutions*. These geometrical solutions are ordered by increasing value of the function cost value (see Figure 29). In return, the topological solutions are ordered with the same order in the manager of the topological solutions.

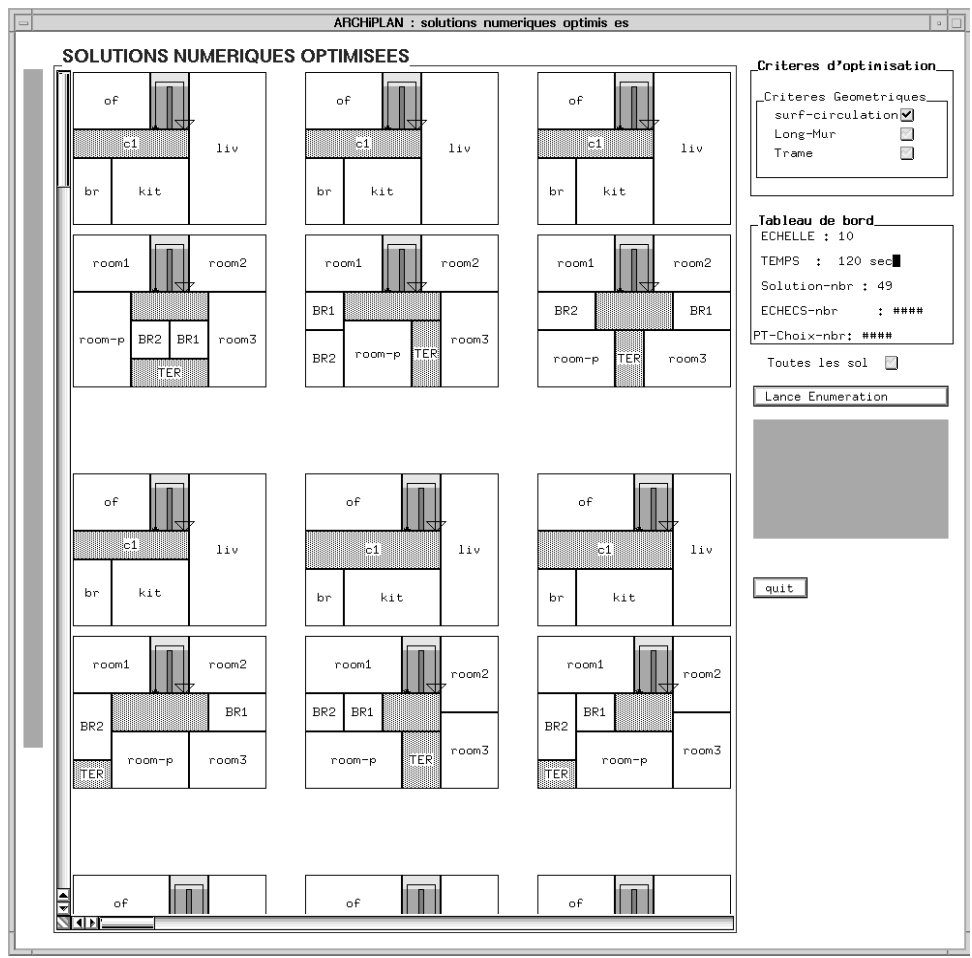

Figure 29 The *n* optimal geometrical solutions, for the *total corridor surface area minimization* criterion, corresponding to the *n* topological solutions of Figure 26.

The architects are not compelled to simply accept numerical optima because the cost function ignores some specific criteria which were very difficult to implement. Anyway, the existing ordering remains somewhat relevant and the architects may benefit from this information to focus on the first topological solutions to achieve an incremental design, as mentioned before. During these incremental designs, the architects could take their subjective preferences into account.

When several global minima exist for a topological solution, a small window with a scrollbar corresponds in its place in the geometrical solutions manager. The scrollbar allows a scrolling on geometrical solutions of minimal cost.

# 6. CASE STUDIES

Several examples in constraint-based space layout planning were tested (Medjdoub, 1996). The results of the classical benchmarks were improved, as for the Maculet (1991) problem (see further). An internal courtyard is allowed. To do this we have two options:

1. The user can relax the "contour total recovery constraint" as we have indicated in section 3.2.1.

2. The user can consider the courtyard as a space. Figure33 shows an example with a Patio. We have put explicitly this patio in the South/East corner, we can put it in the middle of the building

## 6.1 Implementation

ARCHiPLAN has been developed on IBM Risc6000 320H (workstation) in Lelisp v.15 interpreted (object oriented language: Lelisp is a trademark of INRIA), and the constraint library called PECOS (Puget, 1991). The graphic interface has been developed with the AìDA graphic library and Grapher (PECOS, AìDA and GRAPHER are trademarks of ILOG S.A.).

## 6.2 The Maculet problem

The Maculet (1991) problem consists in designing a house with 11 spaces in a building unit contour of 120 m$^2$.

### 6.2.1 Dimensional constraints

*Table 3* presents the dimensional constraints, the module being of 1 meter.

**Table 3** Dimensional constraints for spaces (Maculet problem).

| unit | Area domain value | L-min | W-min | Unit | area domain value | L-min | W-min |
|---|---|---|---|---|---|---|---|
| Floor | [12, 10] | 12 | 10 | Corridor2(c2) | [1, 12] | 3 | 3 |
| Living (sej) | [33, 42] | 4 | 4 | Room1 (ch1) | [11, 15] | 3 | 3 |
| Kitchen (cuis) | [9, 15] | 3 | 3 | Room2 (ch2) | [11, 15] | 3 | 3 |
| Shower (SDB) | [6, 9] | 2 | 2 | Room3 (ch3) | [11, 15] | 3 | 3 |
| Toilet (wc) | [1, 2] | 1 | 1 | Room4 (ch-p) | [15, 20] | 1 | 1 |
| Corridor1(c1) | [1, 12] | 1 | 1 | | | | |

Topological constraints between spaces are :
- *Living* is *On-South-West-contour*,
- *Kitchen* is *On-South-contour* OR *On-North-contour*,
- *Room1* is *On-South-contour* OR *On-North-contour*,

- *Room2* is *On-South-contour* OR *On-North-contour*,
- *Room3* is *On-South-contour* OR *On-North-contour*,
- *Room4* is *On-South-contour*,
- All spaces, except *kitchen*, are *Adjacent* to *Corridor1* OR *Adjacent* to *Corridor2*, with 1 meter minimum for contact length,
- *Living* is *Adjacent* to *Kitchen*,
- *Kitchen* is *Adjacent* to *Shower*,
- *Toilet* is *Adjacent* to *Kitchen* or *Adjacent* to *Shower*,
- *Corridor1* is *Adjacent* to *Corridor2*,
- The contour total recovery constraint is activated (no room is wasted),
- Non-overlapping constraints are activated (spaces don't overlap each other).

In this example, 72 solutions are enumerated in 30 minutes and displayed by ARCHiPLAN (see Figure 30). The 72 corresponding best geometrical solutions, for the corridor surface area minimization criterion, are displayed in Figure 31.

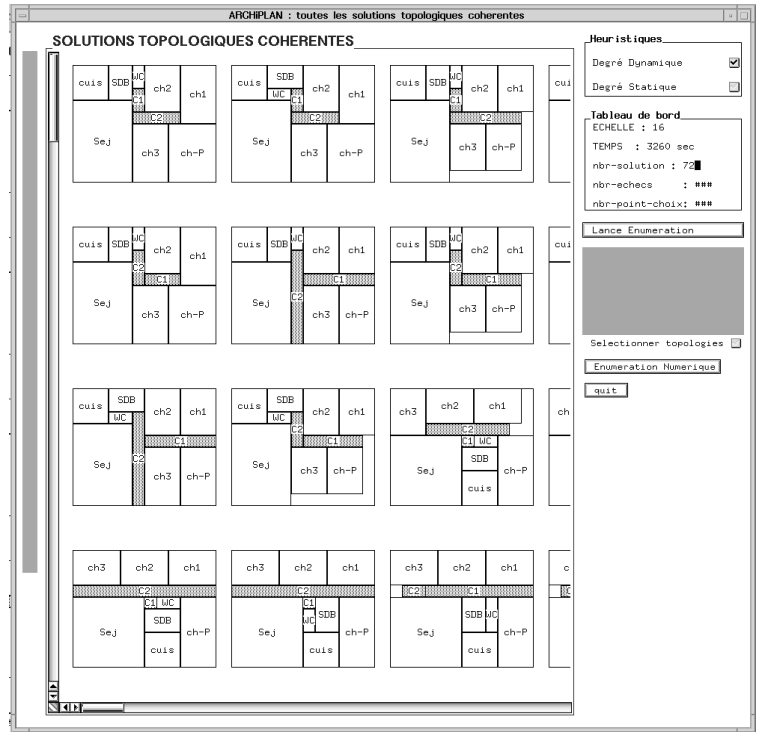

**Figure 30**  Some topological solutions among the 72 possible solutions for the Maculet problem.

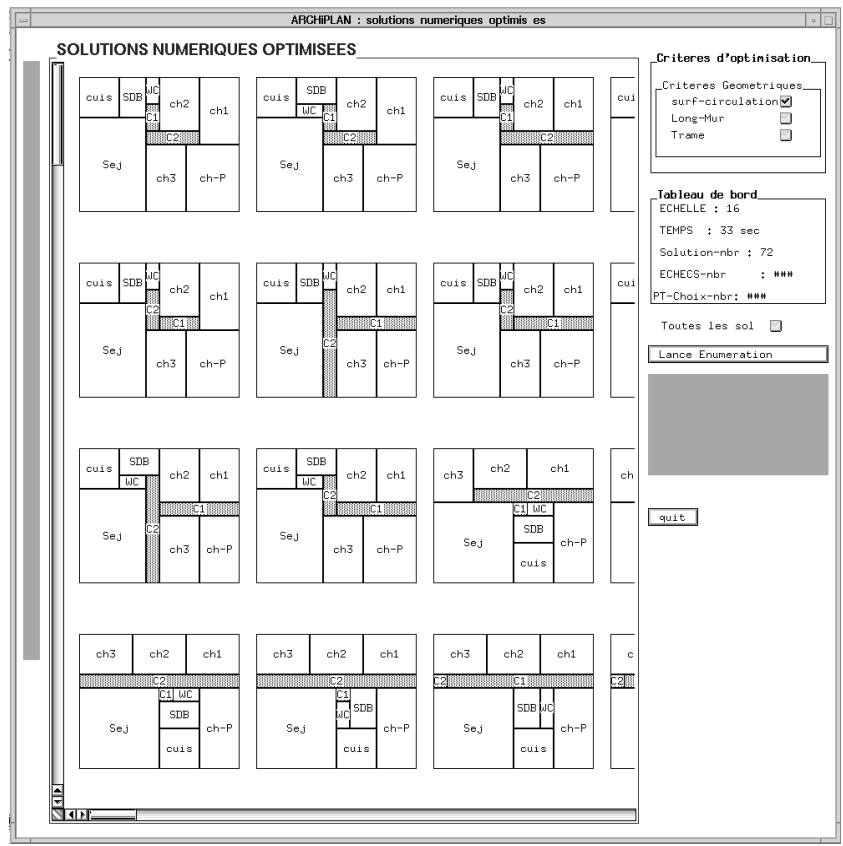

**Figure 31**  Some geometrical solutions among the 72 possible solutions for the corridor surface area minimization criterion.

# 7. CONCLUSIONS and PERSPECTIVES

## 7.1 Our contributions

In this paper, we have presented a model with two solution levels: topological and geometrical, which is close to the methodology of architectural design. Indeed, it has been noticed in architects' opinion that the graphic representation of a topological solution is equivalent to an architectural sketch which has not been seen in other approaches. The partitioning of a relative space orientation into {N,S,W,E} (see Figure 19), the topological solution definition, and topological enumeration heuristics (see [Medjdoub and Yannou, 1998]) are our main contributions. The constraint model has been presented, it is split into *specification constraints* contributing to the functional diagram and *implicit constraints* contributing to reducing the combinatorial complexity. The *generalized adjacency constraint* turns out to be a general and expressive specification constraint thanks to its two arguments of *contact length* and *distance between spaces* (which are also constrained variables). All the constraints have been developed to confer the property of equivalence classes to topological solutions, i.e. one space layout planning (geometrical solution) is adressed by only one topological solution. This way, a topological solution is a conceptual solution.

The ARCHIPLAN approach fits the architect's habits. After the functional diagram definition, ARCHIPLAN automatically generates topological solutions, i.e. all potential sketches without omission. Next, architects may evolve in this topological solution space, which is of rather small size; afterwards, architects can choose some of them for a more detailed study : by incremental design, by an explicit optimization process or by a combination of both. Presently time, determining the best dimensional solution for each topological solution is straightforward because it has already been made for the checking of the topological solution consistency. The current optimization criteria are the total corridor surface area and the total wall length. To finish with, this level of topological solutions limits the combinatorial explosion of geometrical solutions and fits the architect's habits. ARCHIPLAN can be considered as an architect's assistant, able to propose all the conceptual solutions and helping to refine the solutions when the architect's skills are required.

## 7.2 Related works

We have tested numerous examples with ARCHIPLAN (see Medjdoub, 1996). We have improved the results of classical benchmarks (Eastman, 1973; Pfefferkorn, 1975; Tong, 1985). But we have also introduced new benchmarks because a lot of conventional benchmarks in literature seemed to be restricted to simple problems defined by:

- fixed dimensions for building unit contours,
- small number of spaces,
- strongly constrained problems, which is not the case of real problems,
- sometimes spaces of fixed dimensions,
- problems restricted to a unique building unit contour.

ARCHIPLAN is able to cope with all these aspects.

Contrary to the evolutionary approaches (Damski and Gero, 1997; Jo and Gero, 1997) which deal with out-size problems (i.e. Ligett problem, 1985) but obtain under-optimal solutions, our approach deals with middle-size problems (twenty spaces with two floors) with exhaustive enumeration (all the topological solutions) and optimal solutions (one criterion).

We have a complementary approach to the one of (Schwarz et All, 1994) that is based on a graph-theoretical model. In this approach the topological level is apart of the computation process, but the evaluation of the solutions is done at the geometrical level. It is restricted to the small-size problems (doesn't exceed nine rooms) and the shape contour of the building is a result of the design process. In our approach thank to the constraint programming technique and the topological constraints of our model, the variables of the problem are already reduced during the topological enumeration stage. This allows us to represent graphically the topological solutions and, with few effort, to calculate the best corresponding geometrical solutions. The shape contour of the building can be defined before the design process or can be let free with relaxing the *contour total recovery constraint*.

Another feature of our approach is the modular aspect of ARCHiPLAN, which is due to the oriented object programming as well as to the constraint programming (discoupling between constraints and algorithms) which means that the core of ARCHiPLAN will remain unchanged in case of extension of the architectural objects, constraint model or criteria library.

### 7.3 Perspectives

In ARCHiPLAN, many extensions are presently under study:

- Optimization criteria enrichment.
- Generalization of a building unit contour to any shape is essential for this approach to be used for practical problems. The extension of the contour to an assembly of rectangles presents no difficulty. It could even become the object of a user-friendly interface in which the architects would graphically enter the contour.
- Application to the rehabilitation of old buildings (by cost minimization).
- Extension to the industrial space layout planning of a production unit (taking flow constraints of different types into account, and progressive evolution of the production unit).
- Extension to the modelling of more functional specifications according to a primary functional analysis. These functional specifications could be automatically translated to one or several actual ARCHiPLAN's functional diagrams.
- Taking into account the uncertainty of the relevance of a functional specification or its persistences during the life-cycle of the project seems to be a major preoccupation of some industrialists. We therefore envisage to confer to each functional constraint a degree of uncertainty. Functional constraints which are not called into question will be considered as hard constraints and, consequently, they will be submitted to constraint propagation. The other constraints described as "uncertain" will just intervene in the framework of the best geometrical solution research. Each of these constraints will be considered as a criterion of the optimization cost function, the relative importance of this criterion being function of the degree of uncertainty.

## *A.1* **Office Example**

**Table 4** Dimensional constraints between spaces. The dimensions are in a module of 1 meter (L-min stands for *minimum length* and W-min for *minimum width*).

| Unit | Area domain values | L-min | W-min | Unit | Area domain values | L-min | W-min |
|---|---|---|---|---|---|---|---|
| Ft_Floor | [120, 120] | 12 | 10 | Office9 | [9, 15] | 3 | 3 |
| Office1 | [9, 15] | 3 | 3 | Office10 | [9, 15] | 3 | 3 |
| Office2 | [9, 15] | 3 | 3 | Toilet1 | [6, 9] | 2 | 2 |
| Office3 | [9, 15] | 3 | 3 | Toilet2 | [6, 9] | 2 | 2 |
| Office4 | [9, 15] | 3 | 3 | Entrance | [9, 15] | 3 | 3 |
| Office5 | [9, 15] | 3 | 3 | Corridor1 | [1, 30] | 1 | 1 |
| Office6 | [9, 15] | 3 | 3 | Corridor2 | [1, 30] | 1 | 1 |
| Office7 | [9, 15] | 3 | 3 | Patio | [49, 49] | 7 | 7 |
| Office8 | [9, 15] | 3 | 3 | | | | |

The topological constraints are:
- All the spaces are *Adjacent* to *Corridor1* OR *Corridor2*, with 1 meter minimum for contact length,
- The entrance is *Adjacent* to the *building contour*.
- *Corridor1* is *Adjacent* to *Corridor2*,
- The contour total recovery constraint is activated (no room is wasted),

• Non-overlapping constraints are activated (spaces don't overlap each other).

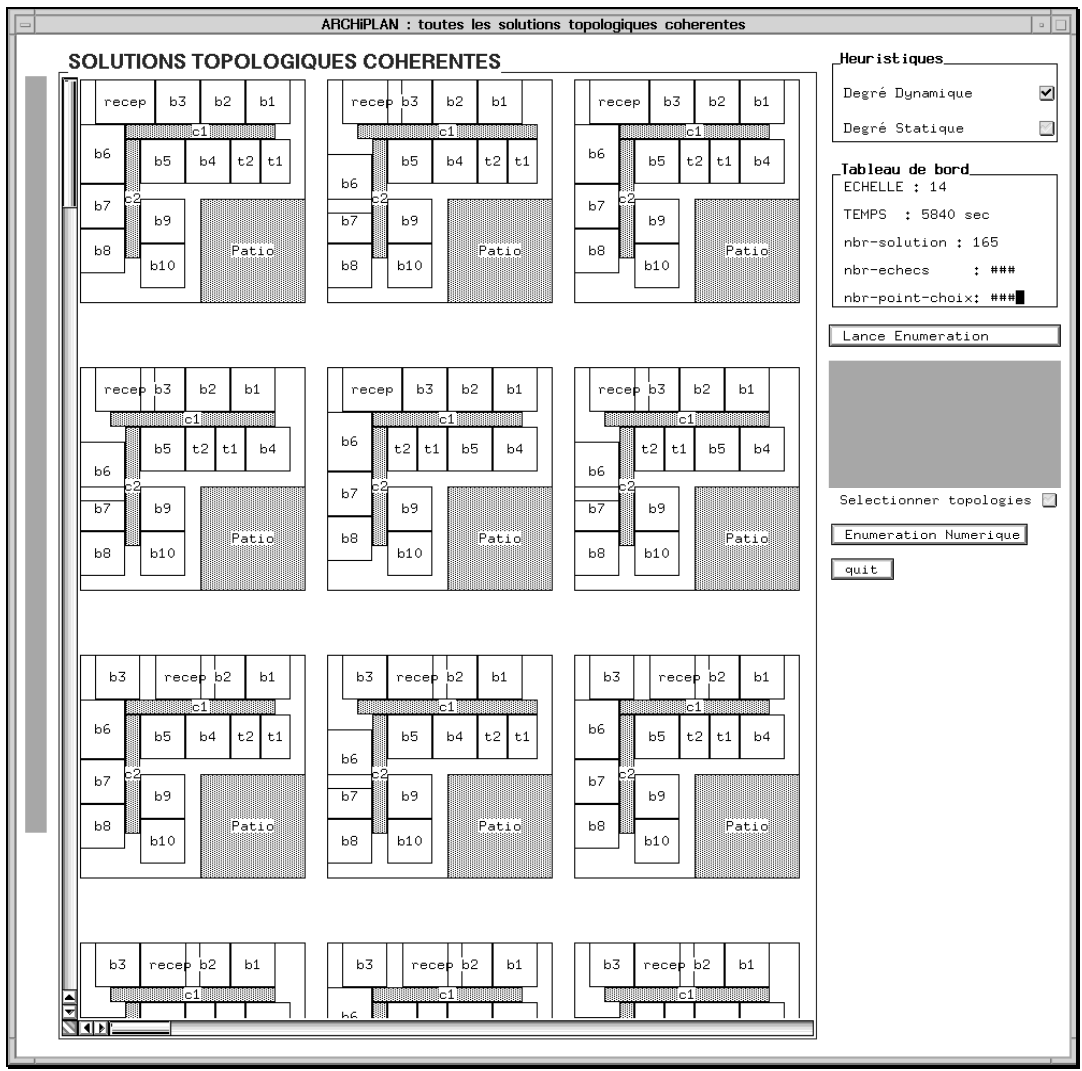

**Figure 32** Some topological solutions among the 102 solutions.

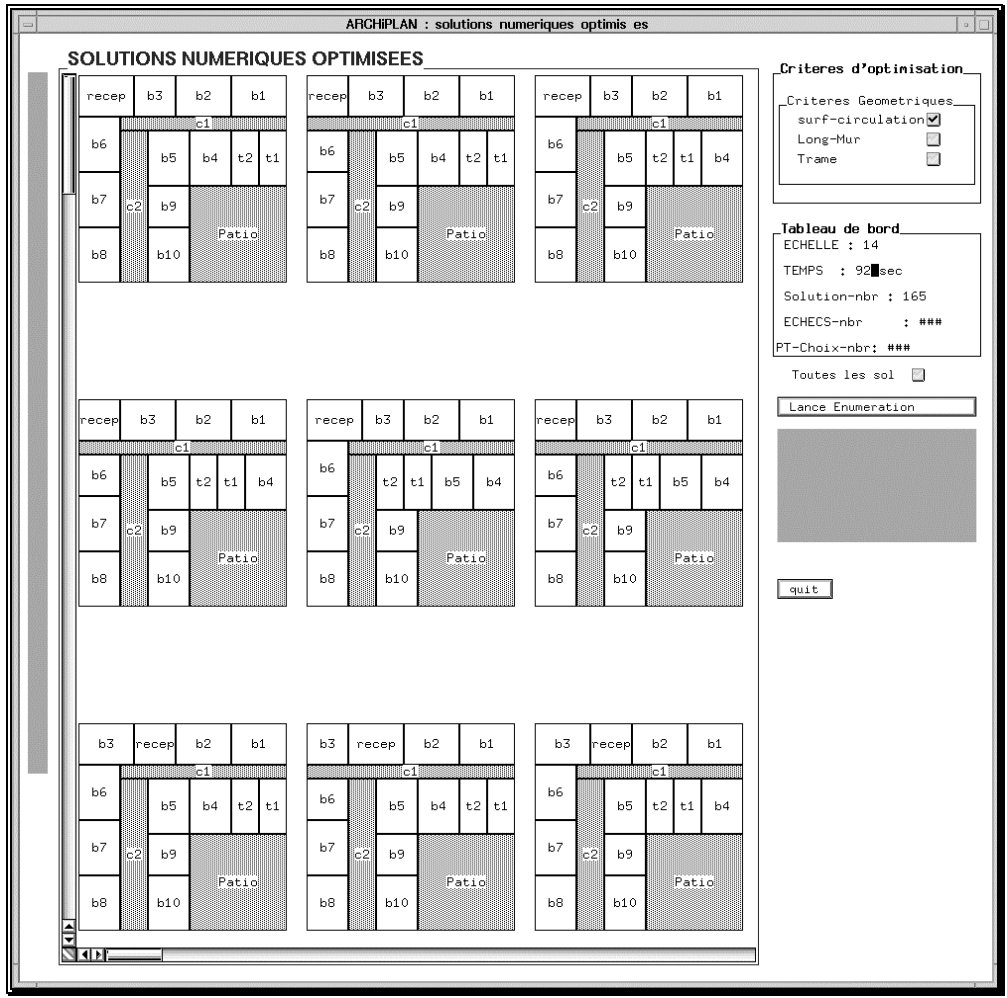

**Figure 33** Some geometrical solutions among the 102 possible solutions with the corridor surface area minimization.